\documentclass[letterpaper, 10 pt, journal, twoside]{ieeetran}  

\IEEEoverridecommandlockouts                              

\usepackage{graphics} 
\usepackage{amsmath} 
\usepackage{amssymb} 
\usepackage{subfiles}
\usepackage{multirow}
\usepackage{makecell}
\usepackage{soul}
\usepackage{color}
\usepackage{wasysym}
\usepackage{colortbl}
\usepackage{xspace}
\usepackage{mathtools}
\DeclarePairedDelimiter{\ceil}{\lceil}{\rceil}
\usepackage{etoolbox}

\newcommand{\naive}{na\"ive\xspace}
\newcommand{\Naive}{Na\"ive\xspace}
    
\usepackage{tikz}
\usepackage{xcolor}
\newcommand*\circled[1]{\tikz[baseline=(char.base)]{
            \node[shape=rectangle,fill,inner sep=1pt] (char) {\textcolor{white}{#1}};}}
            
\usepackage{hyperref}
\usepackage{cleveref}
\usepackage[T1]{fontenc}
\usepackage[linesnumbered, ruled]{algorithm2e}
\DontPrintSemicolon
\SetKwComment{Comment}{\footnotesize$\triangleright\,\,$}{}
\SetInd{0.25em}{0.8em}

\SetCommentSty{mycommfont}

\SetFuncSty{myfuncfont}
\SetAlCapFnt{\footnotesize}
\SetAlCapNameFnt{\footnotesize}

\crefformat{figure}{Figure #1}
\crefformat{section}{\S#2#1#3} 
\crefformat{sections}{\S}
\crefformat{subsection}{\S#2#1#3}
\crefformat{subsubsection}{\S#2#1#3}

\makeatletter
\newcommand{\algorithmfootnote}[2][\footnotesize]{%
  \let\old@algocf@finish\@algocf@finish
  \def\@algocf@finish{\old@algocf@finish
    \leavevmode\rlap{\begin{minipage}{\linewidth}
    #1#2
    \end{minipage}}%
  }%
}
\patchcmd{\algocf@makecaption@ruled}{\hsize}{\textwidth}{}{} 
\patchcmd{\@algocf@start}{-1.5em}{0em}{}{} 
\makeatother

\title{
\textcolor{red}{-- Preprint --} \\
Anytime Planning for End-Effector Trajectory Tracking
}
\author{
Yeping Wang and Michael Gleicher

\thanks{
This work was supported in part by National Science Foundation under Award 2007436 and in part by the Los Alamos National Laboratory and the Department of Energy.}
\thanks{ Both authors are with the Department of Computer Sciences, University of Wisconsin-Madison, Madison, WI 53706, USA $\hspace*{0.8in}$
{\tt\footnotesize [yeping|gleicher]@cs.wisc.edu}}%
\vspace{-0mm}
}
\begin{document}
\maketitle  
\begin{abstract}
End-effector trajectory tracking algorithms find joint motions that drive robot manipulators to track reference trajectories. 
In practical scenarios, anytime algorithms are preferred for their ability to quickly generate initial motions and continuously refine them over time.
In this paper, we present an algorithmic framework that adapts common graph-based trajectory tracking algorithms to be anytime and enhances their efficiency and effectiveness. 
Our key insight is to identify guide paths that approximately track the reference trajectory and strategically bias sampling toward the guide paths. 
We demonstrate the effectiveness of the proposed framework by restructuring two existing graph-based trajectory tracking algorithms and evaluating the updated algorithms in three experiments.  

\end{abstract}

\begin{IEEEkeywords}
Constrained Motion Planning, Motion and Path Planning
\end{IEEEkeywords}

\vspace{-5mm}
\section{Introduction}
\vspace{-1mm}
\IEEEPARstart{M}{any} applications require a robot manipulator to accurately and smoothly move its end-effector along a specific trajectory. These applications involve welding, sanding, polishing, painting, and additive manufacturing. For redundant manipulators or applications with tolerances, there exist infinitely many possible motions to track a reference end-effector trajectory. Thus, planning algorithms seek to find optimal motions among these possibilities based on criteria such as minimal joint movement \cite{rakita2019stampede}, minimal end-effector error \cite{rakita2019stampede,holladay2019minimizing},  minimal maximum joint velocity \cite{morgan2024cppflow}, or minimal number of reconfigurations, \textit{i.e.}, instances where the robot pauses task execution and restarts from another joint configuration \cite{wang2024iklink,yang2022optimal}. 

To track an end-effector trajectory based on some optimality criteria, common graph-based approaches \cite{rakita2019stampede,morgan2024cppflow,wang2024iklink, Descartes,niyaz2020following}   first \textit{sample} inverse kinematics (IK) solutions for each waypoint on the reference trajectory. These IK solutions serve as vertices to construct a layered graph, where each layer corresponds to a waypoint. Edges are added to link IK solutions in adjacent layers, and these edges' weights are defined according to the optimality criteria. Finally, a graph search method is utilized to find the optimal motion. The success of this approach relies heavily on the density of the sampling to sufficiently discretize the solution space. Existing graph-based trajectory tracking approaches often rely on uniformly dense IK sampling, which leads to long delays in finding initial solutions. In practical scenarios with limited computational resources, anytime algorithms are preferred because they can generate feasible solutions quickly and progressively refine them over time. This flexibility allows anytime algorithms to be stopped at any time, enabling trade-offs between solution quality and computation time.

In this work, we present an anytime algorithmic framework that enhances existing approaches to efficiently and effectively generate robot motions to track reference end-effector trajectories. 
Our framework enhances existing methods by incorporating a heuristic to prioritize samples that are likely to lead to good solutions. 
Our key insight is to identify guide paths that approximately track the reference trajectory and strategically bias sampling toward the guide paths, which progressively densifies the graph.  This approach enables frequent identification of effective solutions within a significantly reduced timeframe. Over time, the graph converges towards the dense graph used in conventional algorithms. 
We show that, in the worst case,  our anytime framework converges to the conventional framework, although in practice its ability to sample strategically usually leads to better results in less time.
The algorithmic framework is independent of specific IK sampling algorithms, optimality criteria, or graph searching algorithms. We provide an open-source implementation of our framework
\footnote{Open-sourced code \url{https://github.com/uwgraphics/IKLink/tree/anytime}}.

The central contribution of this work is a guided anytime algorithmic framework (\Cref{sec: alg_overview,sec: technical_details}). To explain our approach, we first describe the conventional sequential framework and a \naive anytime framework as baselines (\Cref{sec: conventional,sec:naive}).
We evaluated our framework through three experiments (\cref{sec: experiments}). First, we applied it to two end-effector trajectory tracking algorithms, Stampede \cite{rakita2019stampede} and IKLink \cite{wang2024iklink}, originally based on the conventional framework. Our results show that our framework accelerates both algorithms, generating solutions with less computation time than the conventional and \naive frameworks, while matching or exceeding their solution quality.
Additionally, we applied it to semi-constrained trajectory tracking, which expands the solution space by allowing tolerances and requiring more IK samplings. Unlike baseline frameworks, which are inefficient due to \naive sampling, our framework enhances IKLink's efficiency and effectiveness for this task.

\vspace{-3mm}
\section{Related Work}
\vspace{-1mm}

\subsection{End-Effector Trajectory Tracking}
\vspace{-1mm}

The end-effector trajectory tracking problem, also known as path-wise inverse kinematics \cite{rakita2019stampede}, task-space non-revisiting tracking \cite{yang2022optimal}, or task-constrained motion planning \cite{cefalo2013task}, involves various approaches based on input type. 
Specifically, trajectory tracking methods \cite{rakita2019stampede, holladay2019minimizing, wang2024iklink, yang2022optimal, Descartes} take a timestamped trajectory as input, whereas path tracking methods \cite{morgan2024cppflow, niyaz2020following} track a sequence of coordinates without timing.

End-effector trajectory tracking methods can be broadly categorized into two groups: trajectory optimization and graph-based approaches. Trajectory optimization directly optimizes a joint-space trajectory while satisfying constraints. For example, Holladay \textit{et al.} \cite{holladay2016distance} utilize TrajOpt \cite{schulman2014motion}, a trajectory optimization method, to minimize the Fréchet distances between current solutions and a given reference trajectory. Instead of Fréchet distances, Torm \cite{kang2020torm} minimizes the summed Euclidean distance between corresponding waypoints on the current and reference trajectories. 
These optimization methods require an initial trajectory as a starting point, are sensitive to this initial trajectory \cite{yoon2023learning}, and are prone to be stuck in local minima \cite{holladay2016distance}.  
Meanwhile, graph-based approaches \cite{rakita2019stampede, morgan2024cppflow, wang2024iklink, Descartes,niyaz2020following} construct a hierarchical graph where vertices represent inverse kinematics (IK) solutions. A path within the graph defines a robot motion by sequentially connecting the IK solutions. We will review these approaches in \cref{sec: conventional} as part of the discussion on the conventional algorithmic framework. Below, we discuss IK sampling strategies in graph-based approaches.

\vspace{-3mm}
\subsection{IK Sampling in Graph-based Trajectory Tracking}
In graph-based end-effector trajectory tracking, the majority of the computation time is consumed by sampling inverse kinematics (IK) solutions and establishing connections between them \cite{malhan2022generation}. 
Some prior work \cite{rakita2019stampede,wang2024iklink, Descartes,niyaz2020following} exhaustively samples IK solutions before searching for a path, resulting in long computation time before obtaining initial solutions. \cite{holladay2019minimizing,malhan2022generation}.
Meanwhile, other approaches use incremental sampling to accelerate the procedure. 
For example, CppFlow \cite{morgan2024cppflow} leverages a generative IK solver, IKFlow \cite{ames2022ikflow}, which rapidly produces a diverse set of IK solutions by learning the distribution of uniformly sampled IK solutions. If the graph search algorithm fails to find a viable motion, the algorithm incrementally samples additional IK solutions to density the graph. 
In contrast to uniform sampling, Malhan \textit{et al.} \cite{malhan2022generation} use continuity in Cartesian space as a heuristic for IK selection during incremental graph construction. However, the heuristic is applicable solely to non-redundant robotic arms. 
Holladay \textit{et al.} \cite{holladay2019minimizing} presents strategies to balance uniform sampling with targeted sampling around certain areas. However, their approach is designed specifically to minimize the Fréchet distance in Cartesian space, and it remains unclear how to adapt their strategies to address objectives such as minimizing joint movements. 
Our method, like Holladay \textit{et al.} \cite{holladay2019minimizing}, balances uniform and targeted sampling, but differs by sampling around a guide path that roughly tracks the reference trajectory.

\vspace{-3mm}
\subsection{Semi-Constrained End-Effector Trajectory Tracking}

In certain trajectory tracking scenarios, precise end-effector movements in all six degrees of freedom are not necessary, \textit{i.e.}, allowing for certain \textit{tolerances} in Cartesian space. For instance, in a drawing task, the robot may be permitted to tilt or rotate the pen as long as the pen tip position remains precise and the motion remains collision-free. This problem is called semi-constrained end-effector trajectory tracking.
Due to tolerances, the solution space expands, requiring more IK solutions to cover the solution space. 
Graph-based approaches with uniform sampling, \textit{e.g.}, Descartes \cite{Descartes}, have a significant increase in computational load when using redundant robots or having tolerances on multiple degrees of freedom \cite{de2017cartesian}. Therefore, developing a more efficient algorithm is essential for semi-constrained end-effector tracking. In this work, we present a guided anytime algorithmic framework to accelerate graph-based trajectory tracking approaches. We show how this framework can efficiently and effectively compute motions to track semi-constrained end-effector trajectory in \cref{sec: experiments}.

\vspace{-1mm}
\section{Technical Overview} \label{sec: technical_overview}
\vspace{-1mm}
In this section, we first formalize our problem statement and describe the conventional algorithmic framework. We then describe a \naive anytime framework and why it is ineffective.
Finally, we give an overview of our guided anytime framework and its advantages over previous approaches. 

\vspace{-3mm}
\subsection{Problem Statement}

Consider a $k$ degree of freedom robot manipulator whose joint configuration and end-effector pose are denoted by $\mathbf{q} {\in} \mathbb{R}^k$ and $\mathbf{p} {\in} SE(3)$, respectively. 
End-effector trajectory tracking refers to the problem of computing a joint space trajectory $\xi{:}[0,T_\chi] {\rightarrow} \mathbb{R}^k$ that drives the robot's end-effector along a desired trajectory $\chi{:}[0,T_\chi] {\rightarrow} SE(3)$, where $T_\chi$ represents the total time duration of the end-effector trajectory. In practice, the target end-effector trajectory $\chi$ is often represented in a discrete form $\{(t_1, \mathbf{p}_1), ..., (t_{n}, \mathbf{p}_{n})\}$, where $\mathbf{p}_i$ is the end-effector pose at timestamp $t_i$. We assume that the waypoints are sufficiently dense to accurately approximate $\chi$. 

In general, there are some requirements for joint space continuity, \textit{e.g.}, in a short time interval $\tau {>} 0$, all joint movements are within their velocity limits, $v_{min} {\leq}[\xi(t{+}\tau)_j$
$ {-} \xi(t)_j]/\tau {\leq} v_{max}, \forall j {\in} \{1,...,k\}, \forall t {\in} [0, T_\chi {-} \tau]$. However, in scenarios where the input trajectory is complex and can not be tracked as a single, continuous path, exceptions to this requirement may be permissible if the task allows for interruptions. In such instances, joint discontinuities may be introduced deliberately. Each joint discontinuity requires the robot to perform an \textit{arm reconfiguration}, in which the robot pauses the task execution, relocates itself in the joint space, and resumes task execution from another configuration. 

This work focuses on the cases where the robot manipulator has some redundancy; it is either a redundant arm ($k>6$) or a non-redundant arm performing semi-constrained trajectory tracking, where certain tolerances are permitted on the end-effector poses. 
Due to the redundancy, the trajectory can be tracked by infinitely many possible motions, requiring optimization criteria to select the best solution.

Graph-based trajectory tracking algorithms maintain a layered graph $G{=}(V, E)$, where each layer corresponds to a waypoint on the reference trajectory. $V$ and $E$ denote vertices and edges in the graph, respectively. The notation $V[x]$ denotes all vertices in the layer corresponding to the $x$-th waypoint. $V[x][y]$ represents the $y$-th IK solution that aligns the end-effector with the $x$-th waypoint. $|V[x]|$ denotes the total number of vertices in $V[x]$.

\begin{figure}[tb]
\includegraphics[width=3.4in]{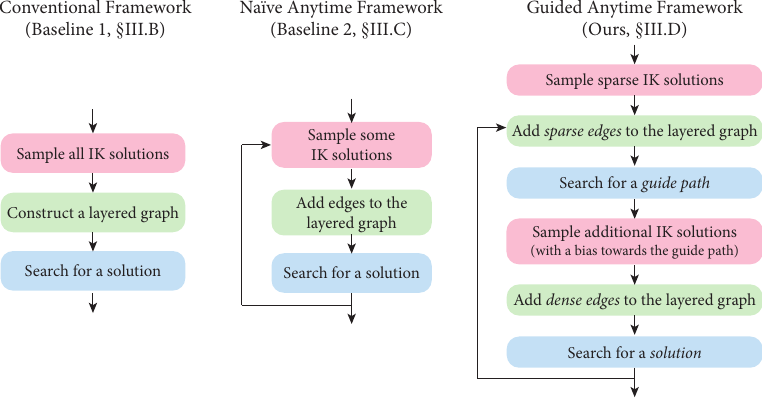}
\caption{Three algorithmic frameworks described and evaluated in this paper} 
\label{fig:methods}
\vspace{-5mm}
\end{figure}

\vspace{-3mm}
\subsection{Baseline 1: Conventional Algorithmic Framework} \label{sec: conventional}

Many existing graph-based trajectory tracking algorithms use a sequential framework. As shown in Figure \ref{fig:methods}-A, these approaches sample inverse kinematics (IK) solutions for each waypoint on the end-effector trajectory, use these IK solutions as graph vertices, and search for a solution in the graph.

\subsubsection{IK Sampling} \label{sec:conventional_sampling}
The two primary goals of IK sampling are to find IK solutions that are \textit{diverse} enough to cover the entire solution space and \textit{dense} enough to construct a smooth motion. 
In order to generate diverse IK solutions, one common approach is to randomly sample starting configurations in a robot's joint space and find IK solutions that are closest to the starting configurations using an optimization-based IK solver, such as Trac-IK \cite{beeson2015trac} or RelaxedIK \cite{rakita2018relaxedik}, or an iterative IK solver, such as Orocos KDL 
or the damped least squares IK algorithm \cite{wampler1986manipulator}. Another approach \cite{morgan2024cppflow} uses a learned generative IK solver, IKFlow \cite{ames2022ikflow}, which learns the distribution of uniformly sampled IK solutions and efficiently generates a diverse set of IK solutions. 

To ensure sufficient density of IK solutions, many previous approaches specify a fixed number of IK solutions for each waypoint, represented by $m$. For example, Stampede \cite{rakita2019stampede} sets $m{=}250$, IKLink \cite{wang2024iklink} sets $m{=}300$, and CppFlow \cite{morgan2024cppflow} sets $m{=}175$. We note that these numbers are used for a 7-DoF robotic arm to track a fully constrained end-effector trajectory. For scenarios involving a more redundant robot (DoF$>$7) or a semi-constrained trajectory, more IK solutions may be necessary. This is because the solution space is larger and requires more IK solutions to densely cover the entire space.

\subsubsection{Graph Construction} \label{sec:conventional_graph}
The sampled IK solutions are used as vertices to construct a layered graph where each layer corresponds to a waypoint on the end-effector trajectory. Edges connect vertices in consecutive layers if the robot can move between the two IK solutions within the time budget specified by the reference trajectory. Edge weights are defined based on various problem formulations such as total joint movement \cite{rakita2019stampede} or maximum joint movement \cite{morgan2024cppflow}.

\subsubsection{Graph Search} \label{sec:conventional_search}
Once the graph is constructed, any paths within the graph represent possible robot motions by connecting the IK solutions. The shortest path from the first layer to the last layer corresponds to a high-quality motion that tracks the trajectory. Different methods are utilized to search for the shortest path. For example, Descartes \cite{Descartes} uses Dijkstra's algorithm, Niyaz \textit{et al.} \cite{niyaz2020following} uses lifelong planning $A^*$, Stampede \cite{rakita2019stampede} uses value iteration, and IKLink \cite{wang2024iklink} uses a dynamic programming method with two objectives.

\textit{Analysis}:
Although multiple IK solutions are sampled for a waypoint, ultimately, only one IK solution is selected to construct the robot motion. Hence, we seek ways to improve the sampling process, \textit{i.e.}, biasing sampling towards areas with a high probability of being selected. Additionally, the number of IK solutions decides the number of edges in the graph; a small set of IK solutions can reduce the time for connectivity checking and graph search.

\vspace{-3mm}
\subsection{Baseline 2: \Naive Anytime Algorithmic Framework} \label{sec:naive}
The aforementioned conventional framework does not provide a solution until completion. 
In contrast, an anytime algorithm provides an initial solution and then iteratively refines it, enabling the process to be stopped at any time and allowing for tradeoffs between solution quality and computation time. 
As shown in Figure \ref{fig:methods}-B,  a \naive anytime framework uses the same random sampling process to initially sample a smaller set of IK solutions, from which a layered graph is constructed. A graph search is then conducted to identify an initial solution.
The solution is iteratively refined by sampling additional IK solutions, expanding the graph, and performing further graph searches to find new solutions. 
The solutions are always at least as good as, if not better than, the previous ones because the graph incorporates all vertices and edges from its previous version. The algorithm continues until it reaches a predefined threshold, such as running time or the number of iterations.

\textit{Analysis}:
The \naive anytime framework has the advantage of generating an initial solution quickly and refining it over time. 
However, it requires some heuristic to find an effective initial solution. In our implementation, we use an optimization-based IK solver to find IK solutions that are close to those of the previous waypoints, which helps ensure smooth motion. This heuristic is similar to GreedyIK in Wang \textit{et al}. \cite{wang2024iklink}.
Despite this, the \naive framework often converges slowly, because the samples are added in a random order. 
This effect is demonstrated in the experiments in \cref{sec: experiments}. 
Meanwhile, our proposed algorithmic framework uses heuristics to bias sampling such that initial solutions are likely to be effective, and the additional sampling focuses on adding samples likely to improve the solution.

\begin{figure*}[tb]
\includegraphics[width=7.0in]{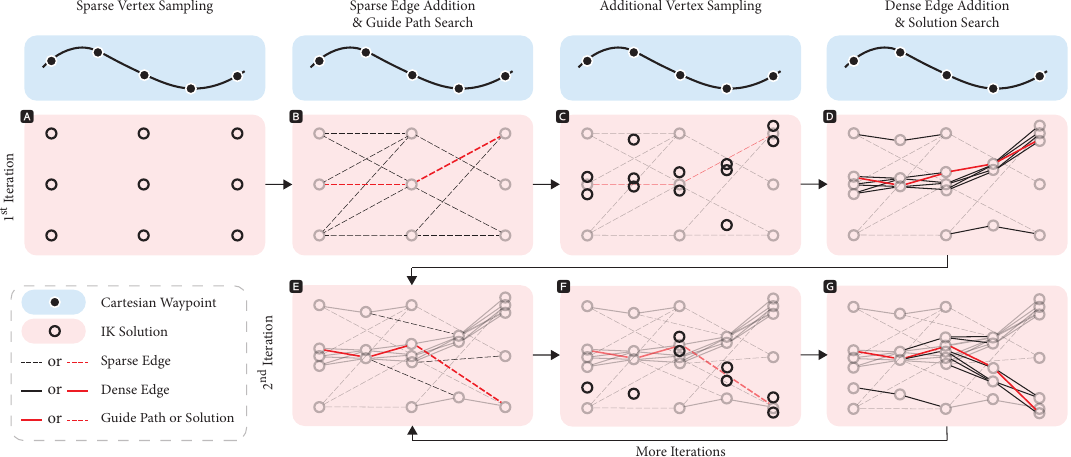}
\vspace{-3mm}
\caption{An illustration of the guided anytime algorithmic framework. In the 1\textsuperscript{st} iteration, \protect\circled{A} a limited set of IK solutions are sampled for specific waypoints. \protect\circled{B} The IK solutions construct a graph with \textit{sparse edges} (dashed lines) and a \textit{guide path} (red) is identified by searching in the graph. \protect\circled{C} Additional IK solutions are sampled with a bias toward the guide path. \protect\circled{D} \textit{Dense edges} (solid lines) are added to connect vertices in adjacent layers and an initial \textit{solution} (red) is identified by searching through the dense edges. In the 2\textsuperscript{nd} iteration, \protect\circled{E} additional sparse edges are added to connect the newly added vertices, and a new guide path (red) is found by searching through both sparse and dense edges. \protect\circled{F} More vertices are sampled, biasing toward the new guide path. \protect\circled{G} After additional dense edges are added, a new solution (red) is identified in the new graph.  In subsequent iterations, the algorithm continues to densify the graph and refine the solutions. } 
\vspace{-5mm}
\label{fig:illustration}
\end{figure*}

\vspace{-3mm}
\subsection{Overview of Guided Anytime Algorithmic Framework} \label{sec: alg_overview}

Similar to the aforementioned approaches, the proposed framework maintains a layered graph, in which a layer corresponds to a waypoint on the reference trajectory and a vertex represents an IK solution. 
The key idea of our method is to bias sampling toward \emph{guide paths}. A guide path is a sequence of vertices that approximately tracks the reference trajectory. It typically includes \emph{sparse edges} that connect vertices that are a few layers apart. Upon refinement, guide paths are likely to be good \textit{solutions}. A solution is a path through the graph with a vertex at every time step, thus accurately tracking the trajectory. During refinement, sparse edges within guide paths are potentially replaced by \textit{dense edges} which connect vertices with adjacent timestamps, thereby gradually converting guide paths to solutions.

As shown in Figure \ref{fig:illustration}, our algorithm starts by constructing a graph in which IK solutions are sampled for certain waypoints and connected using sparse edges. An initial guide path is identified by searching over these sparse edges. The guide path is refined by adding vertices near it and connecting them with dense edges. An initial solution is found by searching over the dense edges. The process is repeated by identifying new guide paths, performing further sampling around them, and finding new solutions. 

\textit{Stages}:
Our framework contains six stages, briefly described here. The details of the stages are outlined in \cref{sec: technical_details}.

\subsubsection{Sparse Vertex Sampling} In contrast to the conventional framework which samples IK solutions for every waypoint, this stage samples a limited number of IK solutions for a subset of waypoints spaced apart. We sample at every $s$ waypoint, where the sparse step size $s$ is a hyperparameter.

\subsubsection{Sparse Edge Addition} This stage adds sparse edges to the graph, connecting vertices that are $s$ layers apart. While sparse edges still correspond to feasible motion segments, they typically do not accurately track the reference trajectory because intermediate waypoints are omitted.

\subsubsection{Guide Path Search} 
This stage identifies a guide path by finding the shortest path from the first to the last layer, using both dense and sparse edges. Guide paths correspond to motions that approximately track the reference end-effector trajectory. The heuristic assumes that high-quality solutions are likely to be found near guide paths. 

\subsubsection{Additional Vertex Sampling}
This stage adds vertices to the graph through a combination of random sampling and targeted sampling around a guide path. The underlying assumption is that IK solutions near the guide path are more likely to yield high-quality solutions. In addition, we also randomly sample an equivalent number of IK solutions. 
This helps avoid local minima where guide paths are misleading and ensures that the algorithm converges to the same worst-case expected performance as the conventional approach.

\subsubsection{Dense Edge Addition}
This stage adds dense edges to the graph to connect vertices in adjacent layers. In addition, certain sparse edges are superseded by the dense edges. 

\subsubsection{Solution Search} 
This stage identifies a solution by finding the shortest path from the first to the last layer using only dense edges. The solution, containing only dense edges, precisely tracks the reference trajectory.

\textit{Incremental Improvement}:
As shown in Algorithm \ref{alg:proposed}, after generating a solution, the algorithm repeats the process, which involves identifying a new guide path, sampling additional vertices with a bias toward this new guide path, adding dense edges to connect the new vertices, and finding a new solution in the new graph. Because each new graph contains vertices and dense edges from its predecessor, each new solution is at least as good as, if not better than, the solution before, thus ensuring incremental improvement. The algorithm terminates upon reaching a predefined threshold, such as maximum running time or sampling capacity.

\textit{Analysis}:
The proposed framework involves random sampling IK solutions, which mimics the process of the baseline approaches. Therefore, given sufficient time, our method will generate the same (probabilistic) sample set as the baseline approaches.
With finite time and sampling budget, even the baseline approaches cannot provide guarantees of optimality, although Holladay \cite{holladay2019minimizing} has proven that a similar algorithm converges towards optimality as the sampling density grows.
Most prior works \cite{rakita2019stampede,wang2024iklink,morgan2024cppflow} empirically show the effectiveness of their algorithms. Similarly, we empirically demonstrate that our sampling strategy often more quickly finds solutions with better or equal quality in \cref{sec: experiments}.

\vspace{-1mm}
\section{Technical Details} \label{sec: technical_details}
\vspace{-1mm}
In \cref{sec: alg_overview},  we gave an overview of our anytime algorithmic framework; this section details its stages.

\begin{algorithm}[tb]
    \footnotesize 
    \caption{\footnotesize Guided Algorithmic Framework}
    \label{alg:proposed}
    \SetKwProg{Fn}{Function}{}{}
    \SetKwProg{For}{for}{\ do}{}
    \SetKwProg{While}{while}{\ do}{}
    \SetKwProg{ForEach}{foreach}{\ do}{}
    \SetKwProg{If}{if}{\ then}{}
    \SetKwInOut{Input}{input}
    \SetKwInOut{Output}{output}
    \SetKwInOut{Hyper}{hparam}
    \SetKwInOut{Global}{g\_var}
    \SetKw{KwTo}{to}
    \SetKw{KwStep}{step}
    \SetKw{KwOr}{or}
    \SetKw{KwIn}{in}
    \SetKw{KwNotIn}{not in}
    \SetKw{KwAnd}{and}
    \SetFuncSty{texttt}
    \SetKwFunction{FDP}{DP}%
    \SetKwFunction{SEARCHH}{SearchHybrid}%
    \SetKwFunction{SEARCHD}{SearchDense}%
    
    \SetKwFunction{SAMPLING}{SparseSample}%
    \SetKwFunction{DEN}{AdditionalSample}%
    \SetKwFunction{CDG}{AddDenseEdges}%
    \SetKwFunction{CSG}{AddSparseEdges}%

    
    \Input{an end-effector trajectory $\chi{=}\{(t_1, \mathbf{p}_1), ..., (t_{n}, \mathbf{p}_{n})\}$ }
    \Output{a joint space trajectory $p$, which is refined over iteration}
    \Hyper{initial number of IK solutions for each waypoint $m_0$, \\
    sparse step size $s$, 
    max number of iterations $i_{max}$}
    \SetKwInOut{Global}{g\_var}
   
    $V \leftarrow \SAMPLING(\chi, s, m_0)$  \Comment*[r]{Sparse vertex sampling} 
    $E \leftarrow \emptyset,$ $i \leftarrow 0$ \\
    \While {$i < i_{max}$} {
        $i \leftarrow i + 1$ \\
        $E \leftarrow  \CSG(V, E, \chi)$ \Comment*[r]{Sparse edge addition}
        
        $G \leftarrow (V, E)$ \\
        $\xi' \leftarrow $ \SEARCHH(G) \Comment*[r]{Guide path search}
        $V \leftarrow \DEN(V, \chi, \xi')$ \Comment*[r]{Sample more vertices}
        
        $E \leftarrow \CDG(V, E, \chi)$ \Comment*[r]{Dense edge addition} 
        
        $G \leftarrow (V, E)$ \\
        $\xi \leftarrow $ \SEARCHD(G)  \Comment*[r]{Solution search}
    }
    \algorithmfootnote{{\texttt{SparseSampling}($\chi, s, m_0$)} randomly samples $m_0$ IK solutions for every $s$ waypoints along the reference trajectory $\chi$. Specific sampling methods are discussed in \cref{sec:conventional_sampling}. \\
    {\texttt{SearchHybrid}($G$)} finds the shortest path from the first layer to the last layer in $G$, using both dense and sparse edges. Specific search methods are discussed in \cref{sec:conventional_search}.  \\
    {\texttt{SearchDense}($G$)} finds the shortest path from the first layer to the last layer in $G$, using only dense edges.
    \vspace{-2mm}
    }
\end{algorithm}

\begin{algorithm}[tb]
    \caption{\texttt{AddSparseEdges}}
    \label{alg:add_sparse_edges}
    \SetKwProg{Fn}{Function}{}{}
    \SetKwProg{For}{for}{\ do}{}
    \SetKwProg{If}{if}{\ then}{}
    \SetKwInOut{Input}{input}
    \SetKwInOut{Output}{output}
    \SetKwInOut{Global}{g\_var}
    \SetKwInOut{Hyper}{hparam}
    \SetKw{KwTo}{to}
    \SetKw{KwOr}{or}
    \SetKw{KwNot}{not}
    \SetKw{KwAnd}{and}
    \SetKw{KwTrue}{True}
    \SetFuncSty{texttt}
    \SetKwFunction{CC}{Cost}%
    \SetKwFunction{CD}{CostDense}%
    \SetKwFunction{ISC}{Linkable}%

    \footnotesize 
    
    \Input{vertices $V$,
            edges $E$ \\
    \hspace{-2mm} an end-effector trajectory $\chi{=}\{(t_1, \mathbf{p}_1), ..., (t_{n}, \mathbf{p}_{n})\}$}
    \Output{updated edges $E$ }
    \Hyper{sparse step size $s$,
        scaling factor $\eta$}
    \For {$x = s{+}1$ \KwTo $n$} {
        \For {$y_1 = 1$ \KwTo $|V[x{-}s]|$} {
            \For {$y_2 = 1$ \KwTo $|V[x]|$} {
                    \uIf {$\ISC(V[x{-}s][y_1], V[x][y_2], t_x {-} t_{x{-}s})$ \KwAnd $\CC(V[x{-}s][y_1], V[x][y_2])$$* \eta < \CD(V[x{-}s][y_1], V[x][y_2])$} {
                        $e \leftarrow (V[x{-}s][y_1], V[x][y_2])$ \\
                        $e.\text{weight} = \CC(V[x{-}s][y_1], V[x][y_2])$ \\
                        $e.\text{is\_sparse} = \KwTrue$ \\
                        $E \leftarrow E \cup \{e\}$
                    }
            }
        }
    }
    \algorithmfootnote{
    \texttt{Linkable}($V_1, V_2, \Delta t$) returns true if the robot can move between the two given joint configurations within the time duration, without exceeding joint velocity limits.  \\ 
    \texttt{Cost}($V_1, V_2$) returns the cost to move directly between the two joint configurations. As discussed in \cref{sec:conventional_graph}, the cost depends on problem formulation. \\ 
    \texttt{CostDense}($V_1, V_2$) returns the lowest cost that moves between the two joint configurations using dense edges. The lowest cost can be computed using graph search methods. If the two joint configurations are not connected by dense edges, return infinity.   \vspace{-5mm}}  
\end{algorithm}

\subsubsection{Sparse Vertex Sampling} \label{sec: sparse_sampling}

This stage involves sampling $m_0$ IK solutions for every $s$ waypoint, starting from index $1$ and including index $n$.
{A total of $\ceil{n/s + 1} \times m_0$ IK solutions are sampled, where $n$ is the number of waypoints on the reference trajectory and $\ceil{\cdot}$ is a ceiling function that rounds a real number up to the nearest integer.}
The specific sampling methods described in \cref{sec:conventional_sampling} can be used to sample IK solutions. 
This stage generates the initial set of vertices in the graph, which are used to find an initial guide path.

\subsubsection{Sparse Edge Addition}
This stage adds sparse edges to the graph.
Algorithm \ref{alg:add_sparse_edges} outlines the {\small\texttt{AddSparseEdges}} function.
In order to add a sparse edge between vertices that are $s$ layers away, the {\small\texttt{Linkable}} function checks if the robot can move between the configurations within the time budget.  
In our implementation, {\small\texttt{Linkable}} assumes linear movements in joint space and checks if a linear movement is within the robot's joint velocity limit. 
In addition, the algorithm checks whether the two vertices are already connected by dense edges. If a path composed of dense edges connects the two vertices and its cost is similar to the cost of the sparse edge, it suggests that the area has already been sufficiently sampled, eliminating the need for a sparse edge.

\subsubsection{Guide Path Search} A guide path is identified by finding the lowest cost path from the first to the last layer of the graph, using both dense and sparse edges. Specific graph search algorithms discussed in \cref{sec:conventional_search} can be used to find the path with the lowest cost.

\subsubsection{Additional Vertex Sampling}

This stage takes a guide path as input and biases sampling toward it. As shown in Algorithm \ref{alg:densification}, this stage first samples around every sparse edge within the guide path, then randomly samples an equivalent amount of IK solutions. 
Sampling around the sparse edges allows them to be replaced by dense edges.

In order to sample around a sparse edge,  we linearly interpolate joint configurations on the sparse edge ({\small \texttt{LinInterp}}), apply some random disturbance ({\small\texttt{AddRandDelta}}), and use it as starting configurations of an IK solver. The IK solver finds IK solutions near the starting configuration. We use RelaxedIK \cite{rakita2018relaxedik} in our implementation, but any other IK solvers mentioned in \cref{sec:conventional_sampling} are also compatible. 

In order to randomly sample IK solutions across the entire trajectory, we first randomly select waypoints using Softmax, which assigns higher selection probabilities to waypoints with fewer randomly sampled IK solutions.
 \begin{equation} 
   \label{eq:probability}
     p_x{=}\frac{\exp(-|V_r[x]|)}{\sum_{i=0}^{n} \exp(-|V_r[i]|)}
\end{equation}
where $x$ is the waypoint index, and $|V_r[x]|$ is the number of existing randomly sampled IK solutions for waypoint $x$. After selecting a waypoint, we generate a random IK solution by initiating an IK solver's starting configuration at a random location ({\small\texttt{RandConfig}}) in the robot's joint space.

\subsubsection{Dense Edge Addition}

This stage adds dense edges to the graph, which connect vertices in adjacent layers. As detailed in Algorithm \ref{alg:add_dense}, the procedure involves iterating over all pairs of vertices located on adjacent layers. The algorithm uses the {\small\texttt{Linkable}} function to check the feasibility of each pair. After adding dense edges, the algorithm identifies and removes the sparse edges that have been superseded by dense edges. Specifically, if there exists a path composed of dense edges that connects the two endpoint vertices of a sparse edge, and the path's cost is similar to the sparse edge's cost, maintaining the sparse edge becomes unnecessary. Our approach shares a similar idea with lazy planning \cite{bohlin2000path, haghtalab2018provable}, which delays the evaluation of edges until they are necessary for the solution. In our approach, sparse edges serve as estimators and a more accurate cost is computed when replacing sparse edges with dense edges. 

\subsubsection{Solution Search} This stage identifies a solution by finding the lowest cost path from the first to the last layer, using only dense edges. Specific graph search approaches described in \cref{sec:conventional_search} can be used to find the shortest path. The identified solution corresponds to a motion that accurately tracks every waypoint on the reference trajectory.

\setlength{\textfloatsep}{0pt}
\begin{algorithm}[tb]
    \caption{\texttt{AdditionalSample}}
    \label{alg:densification}
    \SetKwProg{Fn}{Function}{}{}
    \SetKwProg{For}{for}{\ do}{}
    \SetKwProg{While}{while}{\ do}{}
    \SetKwProg{ForEach}{foreach}{\ do}{}
    \SetKwProg{If}{if}{\ then}{}
    \SetKwInOut{Input}{input}
    \SetKwInOut{Output}{output}
    \SetKwInOut{Hyper}{hparam}
    \SetKw{KwTo}{to}
    \SetKw{KwStep}{step}
    \SetKw{KwOr}{or}
    \SetKw{KwIn}{in}
    \SetKw{KwIs}{is}
    \SetKw{KwNotIn}{not in}
    \SetKw{KwAnd}{and}
    \SetFuncSty{texttt}
    \SetKwFunction{FDP}{DP}%
    \SetKwFunction{SEARCH}{SEARCH}%
    \SetKwFunction{DSEARCH}{DENSE\_SEARCH}%
    \SetKwFunction{SA}{SAMPLE\_AROUND}%
    \SetKwFunction{SR}{SAMPLE\_RANDOM}%
    \SetKwFunction{CC}{COMPUTE\_COST}%
    
    \SetKwFunction{SAMPLING}{SAMPLING}%
    \SetKwFunction{CDG}{CONSTRUCT\_DENSE\_EDGES}%
    \SetKwFunction{CSG}{CONSTRUCT\_SPARSE\_EDGES}%
    \SetKwFunction{IK}{IK}%
    \SetKwFunction{LI}{LinInterp}%
    \SetKwFunction{SI}{SampleIndex}%
    \SetKwFunction{RD}{AddRandDelta}%
    \SetKwFunction{RC}{RandConfig}%
    \SetKwFunction{CP}{CompProb}%

    \footnotesize 
    
    \Input{vertices $V$, 
    a guide path $\xi'$, \\
    \hspace{-2mm}  an end-effector trajectory $\chi{=}\{(t_1, \mathbf{p}_1), ..., (t_{n}, \mathbf{p}_{n})\}$} 
    \Output{updated vertices $V$ }
    \Hyper{number of new IK solutions per waypoint $m_d$ \\perturbation bound $\delta$}
    \Comment{Sample around the guide path} 
    \ForEach {edge \KwIn $\xi'$} {
        \uIf {edge.\textnormal{is\_sparse}} {
            \For {x= e.\textnormal{start\_idx} \KwTo e.\textnormal{end\_idx} } {
                $r = (x{-}e.\textnormal{start\_idx})/(e.\textnormal{end\_idx}{-}e.\textnormal{start\_idx})$ \\
                $\mathbf{q} = \LI(e.\textnormal{start\_vertex}, e.\textnormal{end\_vertex}, r)$ \\
                \For {$i = 1$ \KwTo $m_d$} {
                    $\mathbf{q} =  \RD(\mathbf{q}, \delta)$ \\ 
                    $v = \IK(\mathbf{p}_x, \mathbf{q})$ \\
                    $V = V \cup \{v\}$ \\
                }
            }   
        }
    } 
    \Comment{Randomly Sample} 
    $ \text{Pr} = \CP(V)$ \\
    \For {$i=1$ \KwTo $|V|$} {
        $x \leftarrow \SI(p)$ \\
        $v = \IK(\mathbf{p}_x, \RC())$ \\
        $V = V \cup \{v\}$ \\
        $\text{Pr} = \CP(V)$ 
    }
    \algorithmfootnote{
    \texttt{LinInterp}($V_1, V_2, r$) returns a joint configuration along the straight line between $V_1$ and $V_2$ based on a specified ratio $r$.  \\ 
    \texttt{AddRandDelta}($\mathbf{q}, \delta$) adds random disturbance $\texttt{rand}(-\delta, \delta)$ to each joint value $q_j,\,\, \forall j\in \{1,...,k\}$, and clips 
$\mathbf{q}$ to the valid joint range.  \\ 
    \texttt{IK}($\mathbf{p}, \mathbf{q}_s$) returns an IK solution that moves the end-effector to $\mathbf{p}$. The IK solution is regularized to be near the starting configuration $\mathbf{q}_s$. \\  
    \texttt{CompProb}($V$) returns the probability of selecting a waypoint using Eq. \ref{eq:probability}.
    \vspace{-4mm}
    }  
\end{algorithm}

\setlength{\textfloatsep}{-10pt}
\setlength{\dbltextfloatsep}{-20pt}
\begin{algorithm}
    \caption{\texttt{AddDenseEdges}}
    \label{alg:add_dense}
    \SetKwProg{Fn}{Function}{}{}
    \SetKwProg{For}{for}{\ do}{}
    \SetKwProg{If}{if}{\ then}{}
    \SetKwInOut{Input}{input}
    \SetKwInOut{Output}{output}
    \SetKwInOut{Global}{g\_var}
    \SetKwInOut{Hyper}{hparam}
    \SetKw{KwTo}{to}
    \SetKw{KwOr}{or}
    \SetKw{KwAnd}{and}
    \SetKw{KwFalse}{False}
    \SetFuncSty{texttt}
    \SetKwFunction{CC}{Cost}%
    \SetKwFunction{ICA}{Linkable}%
    \SetKwFunction{INC}{IsDenselyConnected}%

    \footnotesize 
    
    \Input{vertics $V$,
            edges $E$, \\ 
            \hspace{-2mm} an end-effector trajectory $\chi{=}\{(t_1, \mathbf{p}_1), ..., (t_{n}, \mathbf{p}_{n})\}$}
    \Output{updated Edges $E$ }
    \Hyper{scaling factor $\eta$}
    \For {$x = 2$ \KwTo $n$} {
        \For {$y_2 = 1$ \KwTo $|V[x]|$} {
            \Comment{Add dense edges}
            \For {$y_1 = 1$ \KwTo $|V[x{-}1]|$} {
                $e \leftarrow \{(V[x{-}1][y1], V[x][y2])\}$ \\
                    \uIf {$\ICA(V[x{-}1][y1], V[x][y2], t_x {-} t_{x{-}1})$} {
                        $e.\textnormal{weight} = \CC(V[x{-}1][y1], V[x][y2])$ \\
                        $e.\textnormal{is\_sparse} = \KwFalse$ \\
                        $E \leftarrow E \cup \{e\}$
                    }
            }
            \Comment{Remove sparse edges}
            \uIf {$x > s$} {
                \For {$y_1 = 1$ \KwTo $|V[x{-}s]|$} {
                    $e \leftarrow \{(V[x{-}s][y1], V[x][y2])\}$ \\
                     \uIf {$e \in E$ \KwAnd $\CC(V[x{-}s][y_1], V[x][y_2])$$* \eta <$ $\CD(V[x{-}s][y_1], V[x][y_2])$} {
                        $E \leftarrow E \backslash \{e\}$
                    }
                }
            }
        }
    }
\end{algorithm}

\vspace{-3mm}
\section{Experiments} \label{sec: experiments}
\vspace{-1mm}

In this section, we compare our proposed algorithm framework with the two baseline approaches described in \cref{sec: technical_overview}: the conventional sequential framework and the \naive anytime framework. We conducted three independent experiments in simulation, 
each involves two testbeds. The first testbed tracks 10 randomly generated trajectories, comparing our framework with various parameter settings (step size $s{=}3,5, \text{or } 10$) against the baselines. The second testbed tracks a trajectory 10 times and we report the average, best, and worst motions generated by each method to provide
a comprehensive overview of the performance variations.

Our implementation is based on the open source IKLink library  in Rust. We note that the Rust implementation performs faster than the implementation evaluated in the original paper \cite{wang2024iklink}. 
Our framework used $m_d {=} 5$,  $\delta{=}0.2$, $\eta {=} 1.1$ (across all experiments), $m_0 {=} 50$ (Experiments A$\&$B) and $m_0 {=} 500$ (Experiment C).
All evaluations were performed on a laptop with an Intel i7-11800H 2.30 GHz CPU and 16 GB of RAM.

\begin{figure*}[!tb]
\includegraphics[width=7.0in]{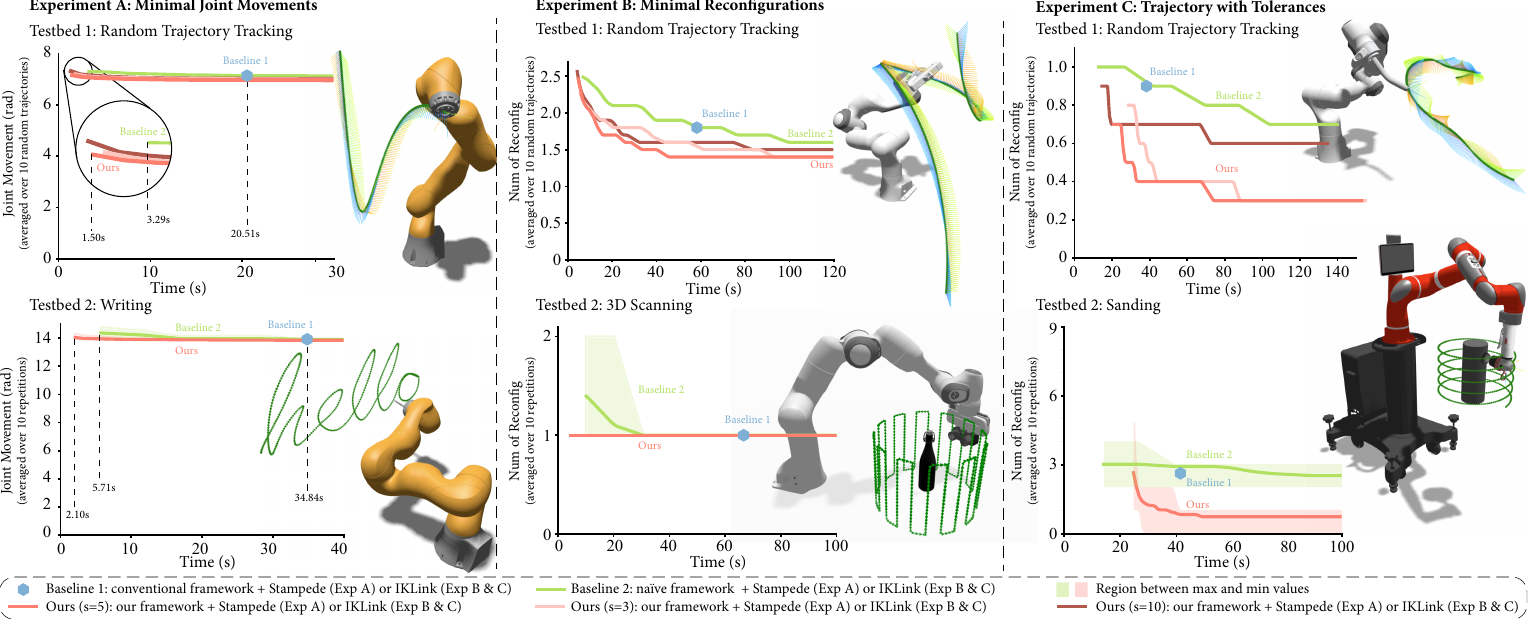}
\vspace{-2mm}
\caption{Results of our three experiments, each on two testbeds. For random trajectory tracking (top row), the results are averaged over 10 randomly generated trajectories. For specific tasks (bottom row), the results are averaged over 10 repetitions of the same trajectory, with colored regions indicating the range between the maximum and minimum values within these repetitions. 
In the line charts, the initial point of a line encodes the average computation time to get initial solutions. 
\textbf{Left column}: our framework enabled Stampede to find initial effective solutions faster than the baselines. \textbf{Middle column}: IKLink with our framework converged faster than with the \naive framework. \textbf{Right column}: for trajectories with tolerances, our framework enabled IKLink to find motion with fewer reconfigurations than the baseline frameworks. The robot visualizations were generated using Motion Comparator \cite{wang2024motion}\protect\footnotemark.
}
\label{fig:results}
\vspace{-3mm}
\end{figure*}

\begin{table*}[tb]
\caption{Experiment Results} 
\vspace{-3mm}
\label{tab:results}
\begin{center}
\begin{tabular}{c|c|c|c|c|c|c|c|c|c|c|c|c}
\hline
\multirow{3}{*}{Method} & \multicolumn{4}{c|}{Exp A} & \multicolumn{4}{c|}{Exp B} & \multicolumn{4}{c}{Exp C} \\
\cline{2-13}
\rule{0pt}{1.05\normalbaselineskip}%
& \multicolumn{2}{c|}{Testbed 1} &  \multicolumn{2}{c|}{Testbed 2} & \multicolumn{2}{c|}{Testbed 1} &  \multicolumn{2}{c|}{Testbed 2} & \multicolumn{2}{c|}{Testbed 1} &  \multicolumn{2}{c}{Testbed 2} \\
\cline{2-13}
\rule{0pt}{1.05\normalbaselineskip}%
& Time$^\dagger$ & JM$^\ddagger$ & Time$^\dagger$ & JM$^\ddagger$ & Time$^\dagger$ & \# RC$^\ddagger$ & Time$^\dagger$ & \# RC$^\ddagger$ & Time$^\dagger$ & \# RC$^\ddagger$$^\ddagger$ & Time$^\dagger$ & \# RC$^\ddagger$
\\  
\hline
Baseline 1\hspace{2ex} & 20.5 & 7.14 & 34.8 & 13.92& 
58.2 & 1.8 & 66.7 & \textbf{1.0} &
38.3 & 0.9 & 41.6 & 2.6\\
Baseline 2\hspace{2ex} & \hspace{1ex}7.1 & 7.14 & 34.4 & 13.92& 
58.9 & 1.8 & 31.2 & \textbf{1.0} &
41.9 & 0.9 & 74.1 & 2.9\\
Ours ($s{=}5$)\hspace{1ex} & \hspace{1ex}\textbf{1.8} & \textbf{6.97} & \hspace{1ex}\textbf{8.1} & \textbf{13.85} & 
\textbf{13.6} & \textbf{1.3} & \hspace{1ex}\textbf{4.4} & \textbf{1.0} &
20.9 & \textbf{0.4} & \textbf{24.6} & \textbf{0.8}\\
Ours ($s{=}3$)\hspace{1ex} & \hspace{1ex}2.5 & 6.99 & / & / & 
20.6 & 1.5 & / & / &
28.7 & 0.6 & / & /\\
Ours ($s{=}10$) & \hspace{1ex}3.5 & 7.01 & / & / & 
22.2 & 1.6 & / & / &
\textbf{14.6} & 0.7 & / & / \\
\hline

\hline
\multicolumn{13}{l}{\rule{0pt}{1\normalbaselineskip}%
$\dagger$: Average computation time to achieve the same performance of Baseline 1 (the conventional framework)} \\
\multicolumn{13}{l}{
\makecell[l]{$\ddagger$: Average performance achieved within Baseline 1's computation time (JM=Joint Movement, \# RC=Number of Reconfigurations) }}
\vspace{-7mm}
\end{tabular}
\end{center}
\end{table*}

\vspace{-3mm}
\subsection{Experiment A - Tracking with Minimal Joint Movements} \label{sec: expA}
We apply our framework to the Stampede algorithm \cite{rakita2019stampede} to find robot motions that track end-effector trajectories with minimal joint movements. 
As proposed in the original paper, Stampede used the conventional sequential framework.
Stampede samples IK solutions by initiating an optimization-based IK solver, RelaxedIK \cite{rakita2018relaxedik}, with starting configurations sampled from a uniform distribution. 
Additionally, it uses value iteration for graph search. 
 Since our objective is to minimize joint movements, we define the edge weights within the graph as the Euclidean distances in joint space between two joint configurations. 

This experiment uses a 7-degrees-of-freedom KUKA LBR iiwa robot. The first testbed tracks 10 random cumulative cubic Bézier curves \cite{kim1995general} for smooth position and orientation changes. These trajectories have an average length of 1.26 m, an average angular displacement of 5.13 rad, and an average of 379.4 waypoints.
 The second testbed replicates a writing task from prior work \cite{morgan2024cppflow,kang2020torm}, where the robot traces a ``hello'' trajectory. The trajectory is rescaled to fit iiwa's workspace. 

As shown in the left column of Figure \ref{fig:results}, while all three approaches generated motions with similar joint movements, our framework achieved initial solutions 2$\times$ faster. The results show that our heuristic is effective for simple trajectories and enables fast identification of effective motions.

\vspace{-3mm}
\subsection{Experiment B - Tracking with Minimal Reconfigurations} \label{sec: expB}

We apply our framework to the IKLink algorithm \cite{wang2024iklink} which utilizes the conventional sequential framework to track the end-effector trajectory of any complexity.  If a reference trajectory is too complex to be tracked as a single, continuous path, IKLink generates motions with minimal reconfigurations. In addition, IKLink has a secondary objective of minimizing joint movements. Similar to Stampede, IKLink uses RelaxedIK \cite{rakita2018relaxedik} to uniformly sample IK solutions. It also leverages RelaxedIK's velocity regularization to find IK solutions close to those in the previous layer, thereby generating smooth motions. Moreover, similar IK solutions are merged using a clustering method. IKLink finds motions using dynamic programming, which incorporates both the number of reconfigurations and joint movements as objectives. 

This experiment uses a 7-DoF Franka Panda robot. The first testbed tracks random trajectories that consist of two consecutive cubic cumulative Bézier curves \cite{kim1995general}. The trajectories average 2.26 m in length, 12.07 rad in angular displacement, and 677.6 waypoints.
The second testbed involves a 3D scanning task, in which a robot accurately moves a sensor along a predefined trajectory, ensuring the sensor consistently points at the object.

As shown in Figure \ref{fig:results}, our framework was more efficient and converged to motions with fewer or equal configurations, compared to both the conventional framework (baseline 1) and the \naive anytime framework (baseline 2).

\vspace{-3mm}
\subsection{Experiment C - Tracking Trajectories with Tolerances} \label{sec: expC}
In this experiment, we show how our framework extends IKLink's capability \cite{wang2024iklink} to track \textit{semi-constrained} end-effector trajectories. 
IKLink was originally designed for fully-constrained trajectory tracking, which requires all 6 DoF of the end-effector to be accurate. This requirement becomes overly restrictive in applications such as welding, where the torch may rotate around its principal axis. In contrast, semi-constrained trajectory tracking allows the end-effector to move within certain tolerances. 
Semi-constrained tracking poses a greater challenge to graph-based approaches because the solution space is larger and requires more samplings to sufficiently cover the solution space. 
We demonstrate that the IKLink using the conventional framework struggles to quickly find plausible motions, and the proposed anytime framework enables IKLinks to efficiently and effectively track trajectory with tolerances. We use RangedIK \cite{wang2023rangedik} as the IK solver, which is an extension of RelaxedIK that generates IK solutions that move the end-effector within specified tolerances. 

The first testbed randomly generates 10 welding trajectories for a Franka Panda robot. Each trajectory is composed of four cumulative cubic Bézier curves \cite{kim1995general}. The trajectories average 1.29 m in length, 22.70 rad in angular displacement, and 646.8 waypoints. Compared to the random trajectories in Experiments A and B, these trajectories involve more rotational movements. In the welding task, the tolerance is along the torch's principal axis, \textit{i.e.}, the robot is allowed to freely rotate its end-effector along the principal axis ($z$-axis). As shown in Figure \ref{fig:results}, the principal axis is angled 45 degrees from the robot's last axis. The robot must coordinate the movement of all its joints to exploit the tolerance, in contrast to just rotating the last joint when the principal axis aligns with the last axis. In the second testbed, a 7-degrees-of-freedom Rethink Robotics Sawyer robot performs a sanding task by tracking a spiral trajectory. The robot can freely rotate the sanding tool around its principal axis, which is oriented 90 degrees from the robot's last axis. 

As shown in Figure \ref{fig:results}, our framework enabled IKLink to quickly find motions with fewer reconfigurations than the baseline frameworks, while IKLink with both the conventional and \naive anytime frameworks struggled to find optimal solutions for trajectories with tolerances.

\footnotetext{Visualization tool: \url{https://github.com/uwgraphics/MotionComparator}}

\textit{Summary}:
Across all experiments, our anytime framework requires less computation time to achieve motions of the same quality and reaches equal or better motion quality within the same time frame compared to both the conventional and \naive anytime frameworks. Figure \ref{fig:results} shows that our framework consistently outperforms the baselines (our lines are consistently below or overlapping with the baselines), while Table \ref{tab:results} presents the corresponding quantitative results.

Experiments A and B show that our framework accelerates the performance of both Stampede and IKLink, enabling them to find motions of higher or equal quality in less time, in comparison to both the conventional and \naive anytime frameworks. Experiment C shows that our framework extends IKLink's ability to efficiently generate motions for tracking semi-constrained trajectories.

Our results also show the effects of the hyperparameter step size $s$, which defines the length of sparse edges. A smaller $s$ increases the number of IK samples in Stage 1, sparse vertex sampling, thereby raising computation time for obtaining initial solutions. Meanwhile, a larger $s$ reduces the accuracy of guide paths as cost estimators, leading to slower algorithm convergence. In our experiments, we observe a sweetspot at $s{=}5$.

\vspace{-1mm}
\section{Discussion}
\vspace{-1mm}
This section summarizes the limitations, implications, and conclusions of our work.

\textit{Limitations}:
Our work has several limitations that suggest future research directions. First, our algorithm lacks a quality-based stopping threshold; users can specify the maximum number of iterations, running time, or sampling budget, but the algorithm does not automatically stop upon convergence. Future work should explore the effects of adaptive stopping thresholds such as when the guide path no longer updates with iterations. 
Second, our method uses a heuristic to bias sampling in graph-based approaches. While we provide empirical evidence to demonstrate its ability to efficiently produce high-quality solutions, we cannot guarantee consistent performance across all scenarios. In certain scenarios, the heuristic may be misleading, causing the method to be slower than the naive anytime framework. Future work may provide a more rigorous understanding of the heuristic's effectiveness.
Third, while we expect that our framework generalizes across many graph-based tracking algorithms, we have only demonstrated it on two. 
Finally, the present work does not consider dynamic constraints, focusing solely on the velocity constraints of robot joints. Future work should consider incorporating acceleration constraints and the inertia of the robot to improve the quality of generated motions.

\textit{Implications}:
The guided anytime framework presented in this work enables graph-based approaches to efficiently and effectively find motions that accurately move a manipulator's end-effector along reference trajectories. It quickly generates initial solutions and continues to refine them as needed.
We have demonstrated that it accelerates and improves solutions over prior methods and enables solving problems that a prior method cannot.
The proposed framework is particularly beneficial for scenarios where end-effector trajectories are complex or have specific tolerances. 

\textit{Conclusion:} This paper presented an anytime framework that enables graph-based trajectory tracking algorithms to efficiently and effectively find robot motions. Our framework generates trajectories of equal or better quality in less time than previous approaches. 


%
%
%
\bibliographystyle{IEEEtran}
\bibliography{citation}

\end{document}